\documentclass[twoside,11pt]{article}
\usepackage{pgm}

\usepackage[utf8]{inputenc}
\inputencoding{utf8}

\usepackage{hyperref,color,soul,booktabs,colortbl,multicol,multirow}
\setulcolor{blue}
\newcommand{\BibTeX}{\textsc{B\kern-0.1emi\kern-0.017emb}\kern-0.15em\TeX}

\usepackage{amsmath}
\DeclareMathOperator*{\argmax}{argmax}

\ShortHeadings{Monotonicity in practice}{Plajner and Vomlel}

\begin{document}

\title{Monotonicity in practice of adaptive testing}

\author{\Name{Martin Plajner}$^{1,2}$ \Email{plajner@utia.cas.cz}\\
\and
   \Name{Ji\v{r}\'{i}~Vomlel}$^1$ \Email{vomlel@utia.cas.cz}\\
   \addr $^1$Institute of Information Theory and Automation,\\
Academy of Sciences of the Czech Republic\\
$^2$Faculty of Nuclear Sciences and Physical Engineering,\\
	Czech Technical University\\
}

\maketitle

\begin{abstract}
In our previous work we have shown how Bayesian networks can be used for adaptive testing of student skills. Later, we have taken the advantage of monotonicity restrictions in order to learn models fitting data better. This article provides a synergy between these two phases as it evaluates Bayesian network models used for computerized adaptive testing and learned with a recently proposed monotonicity gradient algorithm. This learning method is compared with another monotone method, the isotonic regression EM algorithm. The quality of methods is empirically evaluated on a large data set of the Czech National Mathematics Exam. Besides advantages of adaptive testing approach we observed also advantageous behavior of monotonic methods, especially for small learning data set sizes. Another novelty of this work is the use of the reliability interval of the score distribution, which is used to predict student's final score and grade. In the experiments we have clearly shown we can shorten the test while keeping its reliability. We have also shown that the monotonicity increases the prediction quality with limited training data sets. The monotone model learned by the gradient method has a lower question prediction quality than unrestricted models but it is better in the main target of this application, which is the student score prediction. It is an important observation that a mere optimization of the model likelihood or the prediction accuracy do not necessarily lead to a model that describes best the student.
\end{abstract}
 
\section{Introduction}
Computerized Adaptive Testing (CAT) is a concept of testing latent student abilities, which allows creating shorter tests, asking fewer questions while obtaining the same level of information. This task is performed by asking each individual student the most informative questions selected based on a student model. In practice, experts often use the Item Response Theory models~(IRT)~\citep{Rasch1960}, which are well explored and have been in use for a long time. We work with Bayesian Networks (BNs) to model students' abilities instead. This approach can be also found, for example, in \citep{Almond1999, Linden2000}.

Over the last few years we addressed different topics from the domain of CAT. We focused mainly on two topics. The first one is the adaptive testing itself and the use of BN models to perform it, see e.g.,~\cite{Plajner2016b}. The second topic concerns the effect of monotonicity restrictions while learning the model, e.g.,~\cite{Plajner2020}. The current article takes the best from both topics and joins them together in a synergy. Here, we use monotone models to perform simulated adaptive tests. For this purpose we use a data set of the Czech Nation Mathematics Exam\footnote{
The test assignment and its solution are accessible in the Czech language at: 
\url{http://www.statnimaturita-matika.cz/wp-content/uploads/matematika-test-zadani-maturita-2015-jaro.pdf
}}. This exam serves as a high school evaluation exam and the final grade from this exam is considered important. In this article we introduce an approach for inferring the final score and for the prediction of the expected grade of a student. We also provide a method for establishing the 95\% confidence interval of the score which does not require a specific distribution assumption. We observe the evolution of the grade prediction quality during the test and the improvement of the confidence interval. We apply monotone methods, namely our proposed restricted gradient~\cite{Plajner2016b} and the isotonic regression EM by~\cite{Masegosa2016}, as well as the standard (non monotone) EM, and the gradient methods to compare with.

IRT assumes that there is a hidden variable of a student's skill. This approach motivated us to use a structured Bayesian network to model student's skills. In this article we show that the choice of model evaluation criteria is critical in order to select the right model for the given task. It depends whether we want to create a model which predicts the vector of student answers the best or a model which model student’s skills the best. The discovery we uncover in this article is that this distinction is also important in the model selection. Sometimes, the best option is to measure the accuracy of answers prediction or the overall fit of data, i.e. likelihood. The reasonable expectation is that when we are able to do this task the best the model would also model the student the best. Nevertheless, as we discuss in the following sections it is not always the case. We can find models which have worse answer prediction accuracy but they better indicate the student skills as it is reflected by the final score/grade obtained in the test. In other words, the model is less certain about the individual's answers but despite that it models the student better.

The article is structured as follows. In Section~2 we go through the necessary notation and describe the models used. Section~3 brings the methodology for student scoring and grading as well as the formulas to evaluate the precision of models. In Section~4 we describe the experimental settings used for the empirical evaluation and results of these experiments are summarized in Section~5. Finally, Section~6 concludes the paper and recollect the main observations and benefits of this paper.

\section{BN Models and Monotonicity}
\subsection{Models and Adaptive Testing}
In our work we focus on computerized adaptive testing and assessing student knowledge and abilities, using Bayesian Networks with a specific structure. The structure is a bipartite network, which consists of a layer of skills and a layer of questions. Skills are parents in our structure and correspond to specific abilities a student may or may not have. Individual states of these skills are interpreted as levels of knowledge. This interpretation is generally difficult as skills are unobserved variables. Having monotonicity constraints in our models, we are able to introduce an ordering of these levels and refer to them as increasing (or decreasing) qualities of skills. Children in the bipartite structure are question nodes, which correspond to particular questions in a test. Levels of these nodes correspond to the points obtained by solving the specific problem (the problem can be divided into sub-problems with different scores). These models are described in further detail in \cite{Plajner2016}.

\label{sec-notation}
\subsection{Notation}
\label{subsec-notation}
\begin{figure}[t]
	\centering
		\includegraphics[width = 0.9\textwidth]{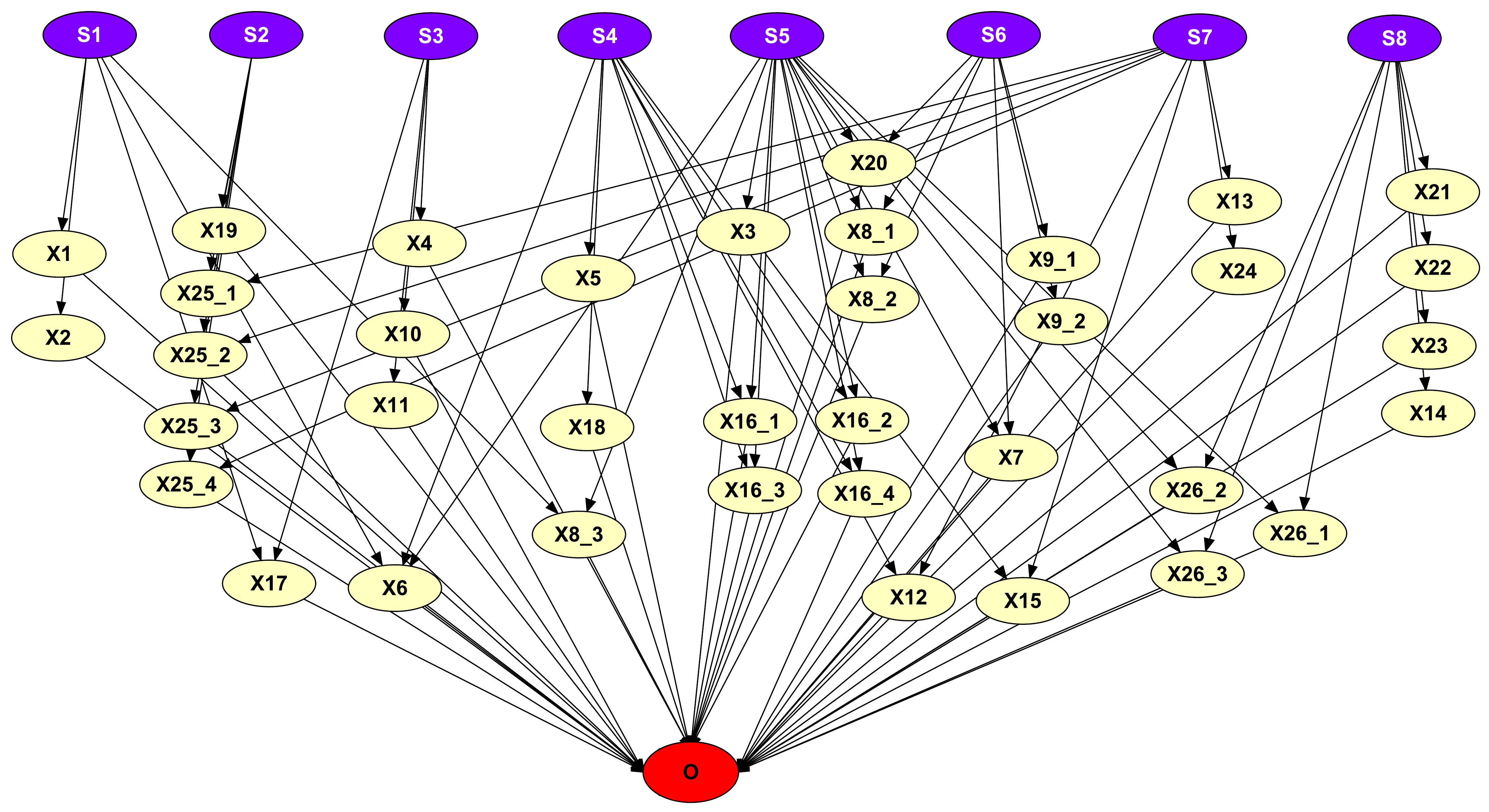}
	\caption{A BN model for CAT}
	\label{fig:CAT_net}
\end{figure}

We use BNs to model students knowledge. Details about BNs can be found, for example, in~\cite{Pearl1988, Jensen2007}. The model we use can be considered a special BN structure such as Multi-dimensional Bayesian Network Classifier which is described, e.g., in~\cite{Gaag2006}. We restrict ourselves to BNs that have two levels of nodes. In compliance with our previous articles, variables in the parent level are skill variables $S$. The child level contains question variables $X$. An example of a BN structure, which we also used in experiments, is shown in Figure~\ref{fig:CAT_net}.

\begin{itemize}
\item We use the symbol $\boldsymbol{X}$ to denote the multivariable $(X_1,\ldots, X_{n})$ taking states $\boldsymbol{x} = (x_1,\ldots, x_{n})$. The total number of question variables is $n$, the set of all indices of question variables is $\boldsymbol{N} = \{1,\ldots,n\}$. Question variables' individual states are $x_{i,t}, t \in \{0,\ldots,n_i\}$ and they are observable. Each question can have a different number of states; the maximum number of states over all variables is $N^{\rm max} = \max\limits_{i}(n_i)+1$. States are integers with the natural ordering. 

\item We use the symbol $\boldsymbol{S}$ to denote the multivariable $(S_1,\ldots, S_{m})$ taking states $\boldsymbol{s} = (s_1,\ldots, s_{m})$. The set of all indices of skill variables is $\boldsymbol{M} = \{1,\ldots,m\}$. Skill variables have a variable number of states, the number of states of a variable ${S_j}$ is $m_j$, and the individual states are $s_{j,k}, k \in \{1,\ldots,m_j\}$. The variable $\boldsymbol{S}^i = \boldsymbol{S}^{pa(i)}$ stands for a multivariable containing the parent variables of the question $X_i$. Indices of these variables are $\boldsymbol{M}^i \subseteq \boldsymbol{M}$. The set of all possible state configurations of $\boldsymbol{S}^i$ is $Val(\boldsymbol{S}^i)$. Skill variables are unobservable.

\item We use the symbol $O$ to denote the score node taking states $o_k, k \in \{ 0,\ldots,\sum_{i \in \boldsymbol{N}}x_{i,n_i}\}$. Its state space is the set of all possible sums of question points and it is modeled as $sum$ rule with questions as parents as shown in Figure~\ref{fig:CAT_net}. The maximum number of points is refered to as $o^m = \sum_{i \in \boldsymbol{N}}x_{i,n_i}$.

\end{itemize}

\subsection{Monotonicity}
The concept of monotonicity in BNs has been discussed in the literature since the 1990s, see~\cite{Wellman1990, Druzdzel1993}. Later, its benefits for BN parameter learning were addressed, for example, by~\cite{Gaag2004, Altendorf2005, Feelders2005}. This topic is still active, see, e.g., \cite{Restificar2013, Masegosa2016}.

We consider only variables with states from $\mathbb{N}_0$ with their natural ordering, i.e., the ordering of states of skill variable $S_j$ for $j \in \boldsymbol{M}$ is
\begin{eqnarray*}
s_{j,1} \prec \ldots \prec s_{j,m_j} \enspace .
\end{eqnarray*}


A variable $S_j$ has an \emph{isotone effect} on its child $X_i$ if for all $k,l \in \{1,\ldots,m_j\}, {t'} \in \{0,\cdots,n_i-1\}$ the following holds\footnote{Note that for $n_i$ this formula always holds since $\sum_{t = 0}^{n_i}P(X_i = t|S_j = s_{j,k}, \boldsymbol{s}) = 1\quad\forall i,\forall j,\forall k$}:
\begin{eqnarray*}
 s_{j,k} \preceq s_{j,l} & \Rightarrow & \sum_{t = 0}^{t'} P(X_i = t|S_j = s_{j,k}, \boldsymbol{s}) \ \geq  \ \sum_{t = 0}^{t'} P(X_i = t|S_j = s_{j,l}, \boldsymbol{s}) 
\end{eqnarray*}
and \emph{antitone effect}:
\begin{eqnarray*}
 s_{j,k} \preceq s_{j,l} & \Rightarrow & \sum_{t = 0}^{t'} P(X_i = t|S_j = s_{j,k}, \boldsymbol{s}) \ \leq  \ \sum_{t = 0}^{t'} P(X_i = t|S_j = s_{j,l}, \boldsymbol{s}) \ ,
\end{eqnarray*}

where $\boldsymbol{s}$ is a configuration of the remaining parents of question $i$ without $S_j$. For each question $X_i, i \in \boldsymbol{M}$ we denote by $\boldsymbol{S}^{i,+}$ the set of parents with an isotone effect and by $\boldsymbol{S}^{i,-}$ the set of parents with an antitone effect. 



Next, we define a partial ordering~$\preceq_i$ on all state configurations of parents $\boldsymbol{S}^i$ of the i-th question, if for all $\boldsymbol{s}^i, \boldsymbol{r}^{i} \in Val(\boldsymbol{S^i})$:
\begin{eqnarray*}
 \boldsymbol{s}^i & \preceq_i & \boldsymbol{r}^{i} \Leftrightarrow
       \left(s^i_j \preceq r^i_j, \ j \in \boldsymbol{S}^{i,+}\right) \ \text{and} \
       \left(r^i_j \preceq s^i_j, \ j \in \boldsymbol{S}^{i,-}\right) \enspace .
\end{eqnarray*}


The monotonicity condition requires that the probability of an incorrect answer is higher for a lower order parent configuration (the chance of a correct answer increases for higher ordered parents' states), i.e., for all $\boldsymbol{s}^i, \boldsymbol{r}^{i} \in Val(\boldsymbol{S^i}),  k  \in  \{0,\ldots,n_i-1\}$:

\begin{eqnarray*}
%
 \boldsymbol{s}^i \preceq_i \boldsymbol{r}^i
& \Rightarrow & \sum_{t = 0}^k P(X_i=t|\boldsymbol{S}^i = \boldsymbol{s}^i)
\ \geq \ 
\sum_{t = 0}^k P(X_i=t|\boldsymbol{S}^i = \boldsymbol{r}^i) 
\enspace .
\end{eqnarray*}

In our experimental part, we consider only the isotone effect of parents on their children. The difference with antitone effects is only in the partial ordering.

%
%

\section{Score prediction and student grading}
In testing it is important to associate a score and/or a grade to a particular student who is being tested. In the adaptive test there are multiple possible options to receive these values. Some options to obtain the student score are described in~\cite{Plajner2016} where, for example, we used estimated skills of the student to compute the score. Nevertheless, the most natural approach seems to be to compute the expected value of the score using the probability distribution of questions' answers in the current state of the student model. In our application we use two different ways how to compute the final score. In both cases we first infer probability distributions of skills of the particular student and then

\begin{enumerate}
	\item[A.] obtain the expected score of remaining unanswered questions, or
	\item[B.] obtain the expected score of all questions (i.e. also those that were already answered).
\end{enumerate} 

Each option is appropriate for a particular scenario. The first one is used in the case the student is tested and we want to estimate his/her result. Questions which were answered define the part of the total score known with certainty and only remaining questions add uncertainty to the total score. This way the test can be evaluated in the just manner. The second approach is more suited for the adaptive learning scenario where we estimate the student score to measure his/her abilities. In this case each question node actually represents a set of similar questions in the test battery. In principal a similar question can be asked again and the answer does not need to be necessary the same, albeit it is most probable it would be. 

\begin{figure}[t]
	\centering
		\includegraphics[width = 0.6\textwidth]{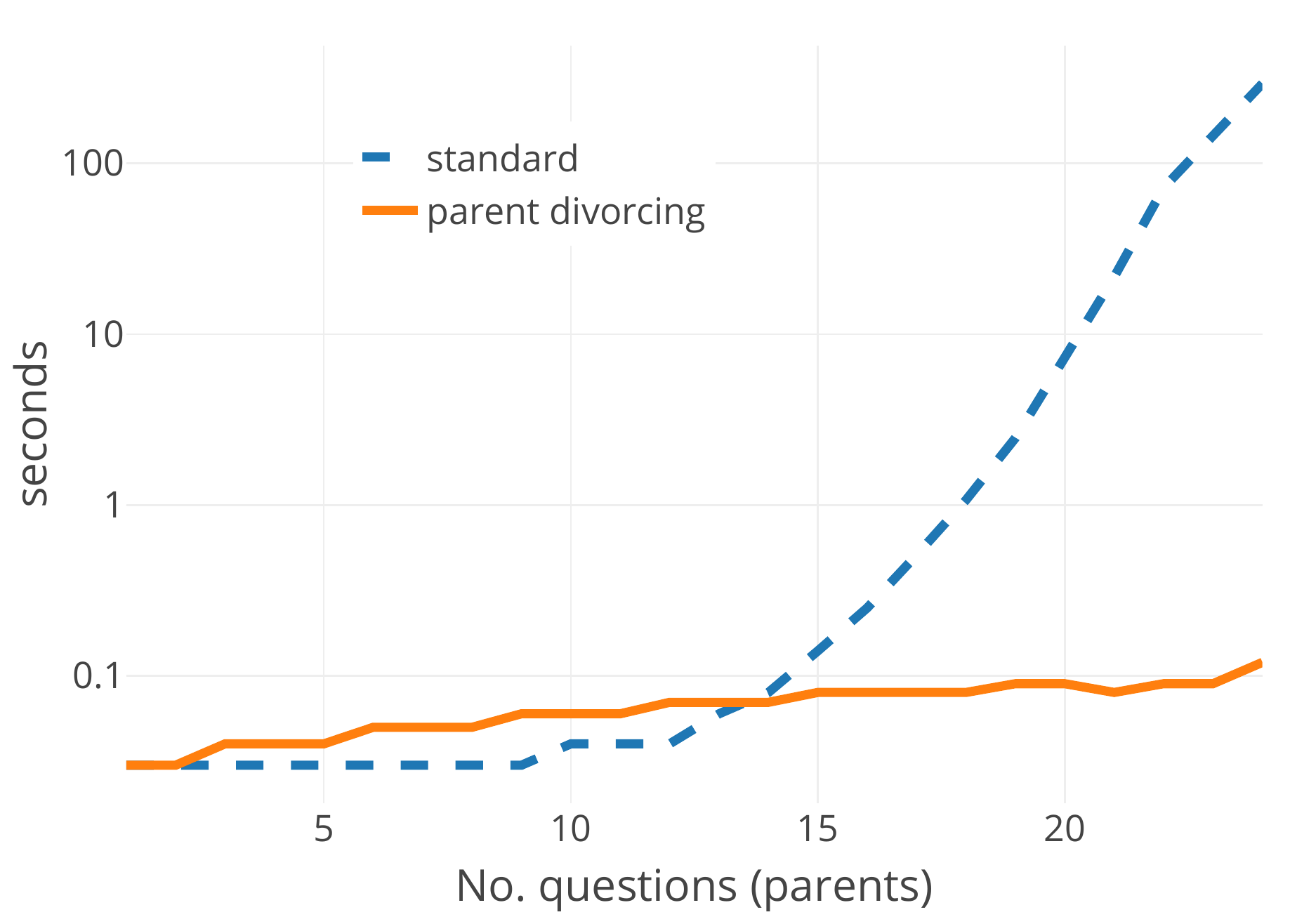}
	\caption{Time of inference in the standard and parent divorcing approach (please note the vertical log scale).}
	\label{fig:parent_divorcing}
	\end{figure}

It is also important to observe not only the expected value but also the distribution of the score. We model the score distribution by an additional node in the Bayesian network, the score node $O$. This node has as many states as there are possible points to be obtained in the test. The node probability distribution is given by a simple $sum$ rule of its parents (questions). The problem which we have to address in this case is that the dimension of the CPT of this node is very large. We work with 37 questions which have two or more states. Even if they were binary the full state space would be of dimension of $2^{37}$. This value is very large and it does not allow direct inference due to the memory size limit. We use the parent divorcing method as described in~\cite{OLESEN1989}. Another option is to use the rank-one decomposition as it is described in~\cite{Savicky2007}. The reduction of the computational time is very significant as it is outlined it in Figure~\ref{fig:parent_divorcing}. We show the increase of the time necessary to perform the inference based on the number of questions we connect together. The computational time of the inference is computed only for smaller number of questions as it is not feasible for the standard case in larger numbers.

In this way we obtain the distribution of student's score over the point scale as shown, for example, in Figure~\ref{fig:adaptive_dist}. Using this distribution we can estimate the expected score and its 95\% confidence interval as well. This confidence interval is obtained in the following manner. We sort the states of the node $O$ (points scale) in terms of the states' probabilities in the descending manner. We select all states until the total cumulative probability exceeds 0.95. The probability distribution over tends to be similar to the Gaussian distribution but it is not a rule. The advantage of the proposed approach is that it does not require a specific distribution assumption.

We establish two key performance measures to evaluate different methods in the adaptive testing scenarios. We measure 
\begin{eqnarray}
\text{the accuracy of answers prediction}  & \enspace & a^Q = \dfrac{\sum_{i \in \boldsymbol{N}}{I(x_i = x_i^*)}}{n} 
\label{eq:aq}
\\
\text{the abs. error of the total score prediction} & \enspace & e^S = |o^* - \sum_{j = 1}^{o^m}j\cdot P(O = j)|
\label{eq:es}
\end{eqnarray}
in each step of the adaptive test for each learning method used, where the function $I(x_i = x_i^*)$ returns one as the maximum likelihood state equals the observed state for the question $i$ and $o^*$ is the real obtained score. 

From the perspective of a student the most important measure is the test grade; especially in our special case of the National Exam. The problem of assigning a grade to a student can be viewed as a classification problem which aims at placing a student into the correct grade category. We assume there are $G$ grades where each grade is given for the resulting score in a range of points $G_i, i \in {1,\ldots,G}$. The expected grade $g$ is then established from the score variable as
\begin{equation}
g = \argmax_{i\in {1,\ldots,G}}(\sum_{j \in G_i}P(O = j)) \enspace .
\label{eq:}
\end{equation}
The error of this classification is then computed as
\begin{equation}
e^g = \sum_{i\in {1,\ldots,G}}|g^*-i|\cdot\sum_{j \in G_i}P(O = j) \enspace ,
\label{eq:grade_err}
\end{equation}
where $g^*$ is the observed grade.

\section{Experimental setup}
For experiments in this article we use the data set of the Czech National Mathematics Exam. This exam is taken at the end of the high school and the same test is taken by each student in the same term. Given the nature of this test these data set is valid in terms of student motivation to complete the test as good as possible and data quality is high.

Our experiments were performed according to the following scheme. From all available tests we first drew a random subset to serve as a training set. With each set we train BN models with different learning methods, namely regular EM, regular gradient (grad), isotonic regression EM (irEM), quick~irEM (qirEM), and restricted gradient (rgrad). There are 10 random starting points, same for each method to start the learning process at. From the resulting 10 learned models we select the winning model based on the optimization criteria which is the log-likelihood value measured on the training sample. In our previous article~\cite{Plajner2020} we compared individual methods on the log-likelihood of the complete data set. In this article the main focus is on adaptive testing usage and we simulate the adaptive testing scenario. The procedure above is performed 10 times with different data selected for learning for each learning set size of 10, 40, and 160 students (i.e. test results). Final ten results for each learning set size are then averaged over the measured metrics.

These learned models are further used in the adaptive testing scenario. We select 100 students which did not figure in any previously selected sets. These students are tested in the simulated test. Tests are performed in two different ways
\begin{itemize}
	\item fixed and
	\item adaptive.
\end{itemize}
The first one is selected in order to provide better insight into comparison of methods. In the adaptive version of testing different questions may be selected for each method in each step. This fact makes the comparison harder in some aspects. On the other hand the ability of a model to be used adaptively is a desired one and we provide comparison of both approaches as well.

\section{Results}

\subsection{Student classification}

\begin{figure}[t]
	\centering
		\includegraphics[width = \textwidth]{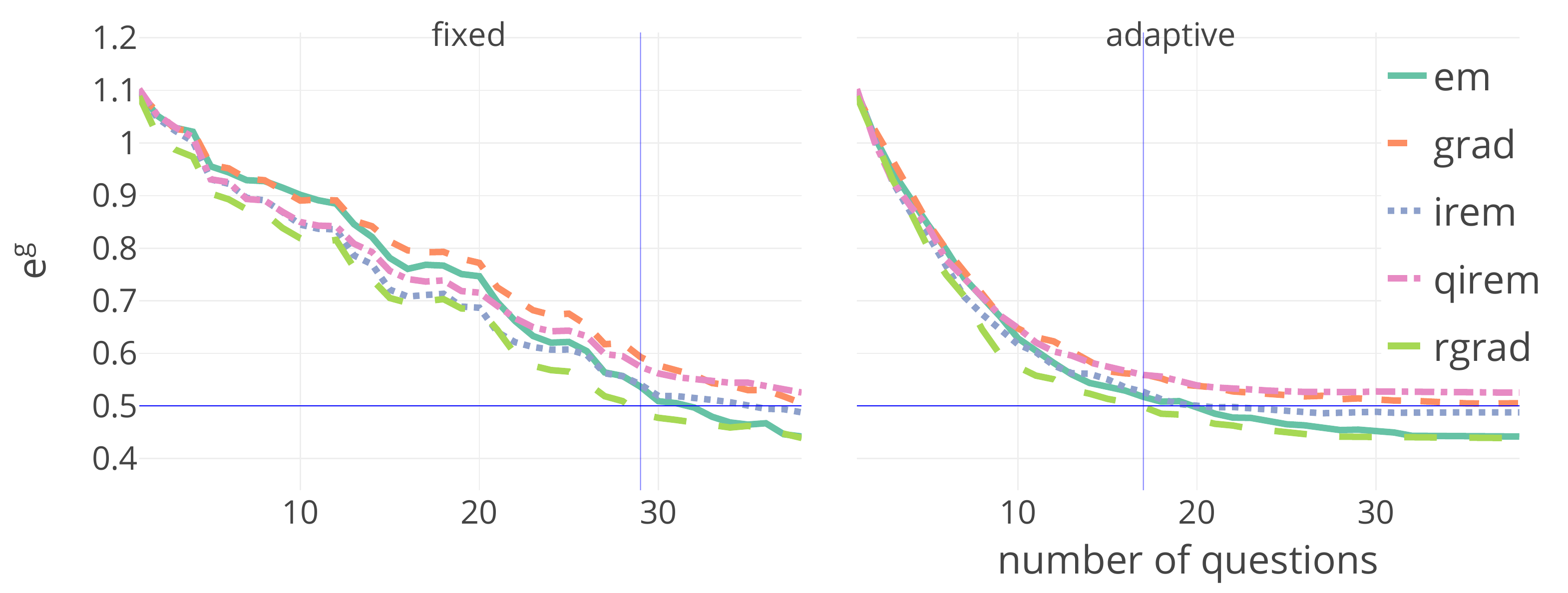}
	\caption{Evolution of the grade prediction error based on skills (B) for models of the learning set size 10, fixed and adaptive question selection.}
	\label{fig:grade_err_mirror}
	\end{figure}

The grading in the Czech National Mathematics Exam is given by the following scheme.\\
0-16 points: 5; 17-25 points: 4; 26-34 points: 3; 35-43 points: 2; 44-52 points: 1\\
In the experiments each student is assigned the expected final grade in every step and the error $e^g$ of this assignment is measured as described in~(\ref{eq:grade_err}). Figure~\ref{fig:grade_err_mirror} shows the evolution of this error during the adaptive test with models with the learning set of size 10. We show only the version B of questions' answers predictions based only on the inferred skill. Because of that it does not end in zero as we never have the absolute certainty of a student score while knowing only his/her skills. This corresponds to the real-life situation with the margin for errors and mistakes during taking the test even by the best students. In this figure we can see that the restricted gradient method provides the best results. In the end of the test it is on the same level as unrestricted EM and better than other methods. Due to the limited space, we do not include the case A of questions' answers where we predict only remaining questions because it behave very similarly in the most important part of the testing, i.e. aprox. the first half. In the second half it converges to zero as we already know all the answers and thus also the final grade is known.

Special attention should be given to the comparison of the fixed and adaptive approach. Notice the horizontal and vertical lines which mark the threshold of passing the error of 0.5 in all cases. This error threshold is passed in the question 29 and 17 for the fixed and adaptive variants respectively. This observation provides several outcomes. The first one is that using the adaptive version of test significantly reduces the number of questions we have to ask in order to obtain the same level of information. By inspecting the adaptive version of the skills variant further we can see that after asking the first 17 adaptive questions we obtain almost as much information about student skills as possible which gives an option of shortening the test.

  %
  %
	%

\subsection{Score and answers prediction}
\begin{figure}[t]
	\centering
		\includegraphics[width = \textwidth]{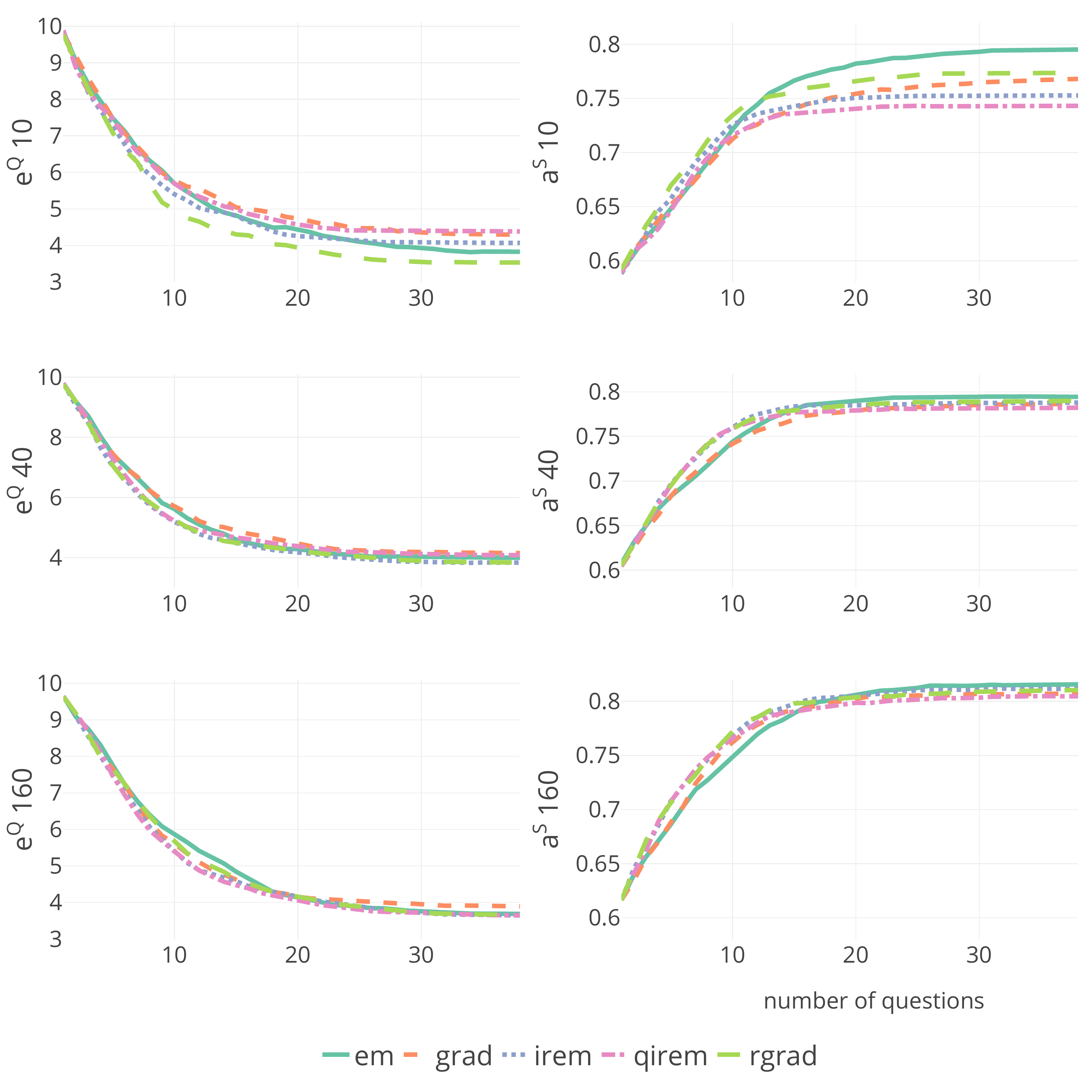}
	\caption{Evolution of the abs. error of the total score prediction $e^S$ and the accuracy of answers prediction $a^S$ for different learning set sizes.}
	\label{fig:S_acc}
	\end{figure}

Figure~\ref{fig:S_acc} shows the measures of the grade prediction error $e^S$ and the answers prediction accuracy $a^Q$ as they are defined in equations~\ref{eq:aq} and~\ref{eq:es}. By inspecting this figure we can see that the restricted gradient method outperforms all other methods in the grade predictions. The only exception where it is slightly worse is the middle part of test for the largest learning set. The highest difference is in the smallest learning set where its benefit is visible the best. In the prediction of answers, restricted gradient method is better in the early stages of testing. For larger learning sets together with other monotone methods (irem and qirem). In the smallest set it is the best of all tested methods. Nevertheless, the best method in the final parts of testing is unrestricted EM. This difference between prediction quality of score and answers is very interesting and it is discussed further in Conclusions section.
	
\begin{figure}[t]
	\centering
		\includegraphics[width = \textwidth]{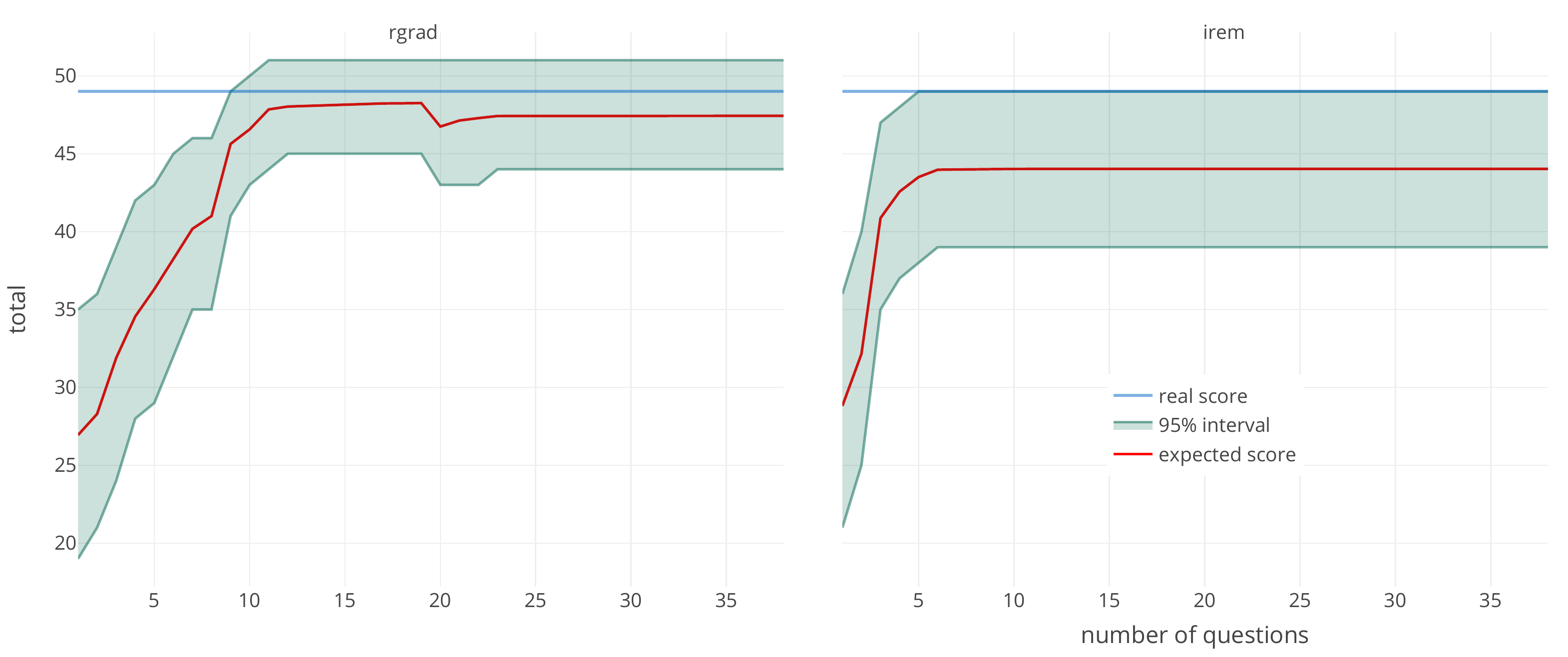}
	\caption{Evolution of the of the total score prediction and its confidence interval for an individual test during the adaptive procedure for the restricted gradient and irem methods, 10 learning samples.}
	\label{fig:adaptive}
	\end{figure}

\begin{figure}[t]
	\centering
		\includegraphics[width = \textwidth]{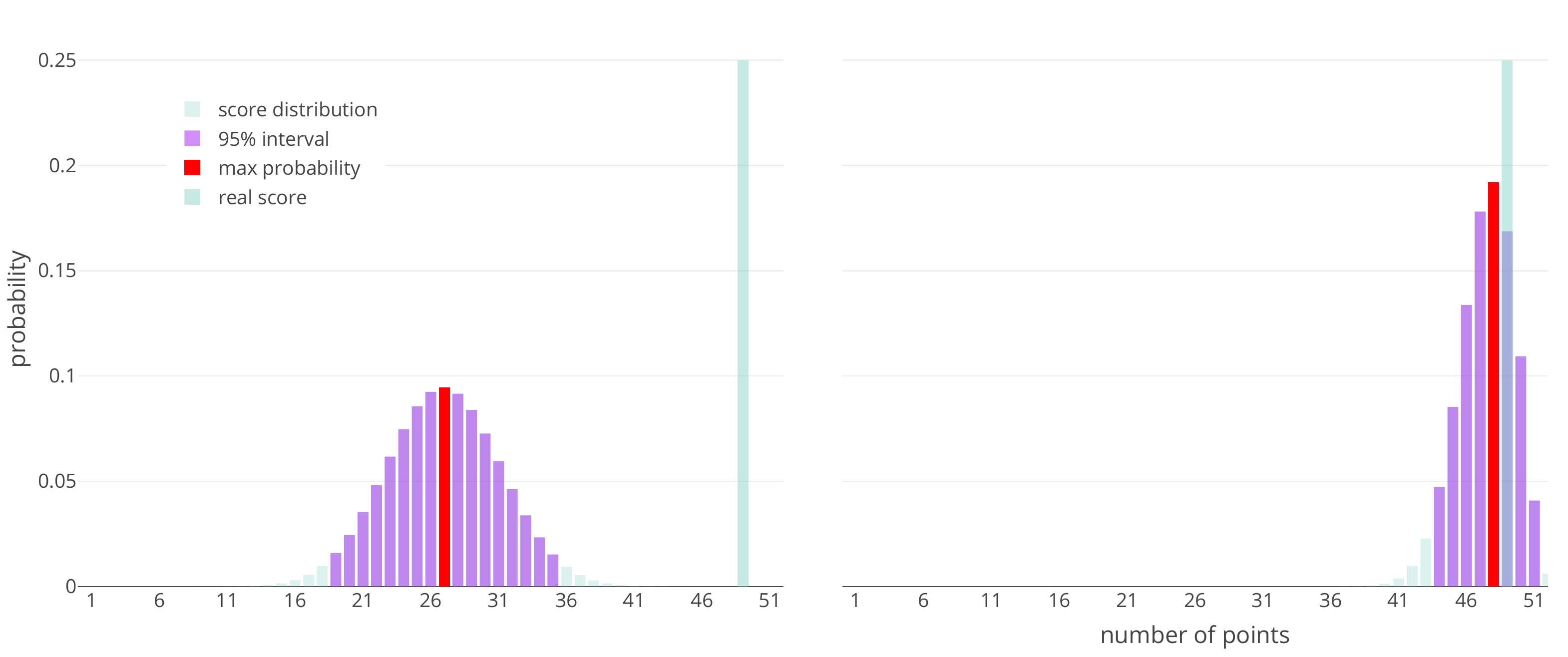}
	\caption{The expected score and its confidence interval for an individual test at the start and the end of testing. Restricted gradient method, 10 learning samples.}
	\label{fig:adaptive_dist}
	\end{figure}

Figures~\ref{fig:adaptive} and~\ref{fig:adaptive_dist} show the evolution of the score prediction as the states of the node $O$ with its confidence interval as it is described in Section 3. Results displayed are obtained from the adaptive test simulation of an individual test with models learned from 10 observations with the rgrad method and the irEM method. The first figure shows the expected value and its confidence interval during the whole test for both methods. We can see that in this particular case both methods shift to the better predictions quite quickly. In this case irEM is faster and at the fifth question its prediction of the total score is better. Nevertheless, irEM stays at the same level for the rest of the test while the rgrad method improves and its final assumption is only approximately one point of the real total score. Another important fact to notice is the shrink of the confidence interval. For the rgrad method it starts at the width of 17 points and it ends at the width of 7 points. This situation is further detailed in Figure~\ref{fig:adaptive_dist} where we show the probability distribution at the start and the end of the same test\footnote{For the sake of visualization simplicity we display the most probable score instead of the expected score in this case.}.

	%

	%
	%
	%
	%
	%

\section{Conclusions}

This article explored the impact of monotonicity restrictions in BN models used to model students in the Czech National Mathematics Exam. It also showed the benefit of adaptivity in testing with this specific data set. In experiments we used monotone and non monotone methods and performed comparisons using different evaluation criteria.

The first observation is the benefit of the adaptive approach to testing. As it can be clearly seen in Figure~\ref{fig:grade_err_mirror} the number of questions we need to ask is reduced by one third. This creates the space either reducing the length of the test, or using the extra time to increase the precision by asking other questions better tailored for the particular student.

Another new aspect we discussed is the prediction of the total score of a student which is an indicator of his/her skills. We proposed a methodology for measuring the score including the corresponding confidence interval. We compared results of monotone methods and we showed the evolution of the score and the confidence interval during the testing.

Last but not least, we would like to emphasize that monotonicity improves the quality of the grade, score, and question answers predictions. Especially, when the learning set is small and at the first stages of testing. Our empirical results show that the restricted gradient method we propose provides the best results of all tested methods at the first stages of the test. At the later stages of the adaptive test, the regular EM algorithm learning method provided models which were the most precise in terms of individual question answers. This is caused by its flexibility in learning. As EM is not restricted by monotonicity it can learn dependencies monotone methods can not and that allows it to model question answers more precisely in some cases. This result is interesting in the context of the score prediction quality which is an observable indicator of the student skills. When this metric is used for the model evaluation, the EM models were outperformed by the restricted gradient models despite the restricted gradient models prediction of individual answers was worse. The reason is that the monotone models are able to better model the student himself/herself. They are not certain about individual questions but they better infer the score since it is based on their skill model which better characterizes the tested student. This observation means that it is important to keep in mind the purpose of a model while learning it. This is a general observation valid not only for CAT but also for other applications.

\appendix
\acks{This work was supported by the Czech Science Foundation (Project No. 19-04579S) and by the Grant Agency of the Czech Technical University in Prague, Grant No. SGS17/198/OHK4/3T/14.}

\vskip 0.2in
\bibliography{bibli}

\begin{thebibliography}{18}
\providecommand{\natexlab}[1]{#1}
\providecommand{\url}[1]{\texttt{#1}}
\expandafter\ifx\csname urlstyle\endcsname\relax
  \providecommand{\doi}[1]{doi: #1}\else
  \providecommand{\doi}{doi: \begingroup \urlstyle{rm}\Url}\fi

\bibitem[Almond and Mislevy(1999)]{Almond1999}
R.~G. Almond and R.~J. Mislevy.
\newblock {Graphical Models and Computerized Adaptive Testing}.
\newblock \emph{Applied Psychological Measurement}, 23\penalty0 (3):\penalty0
  223--237, 1999.

\bibitem[Altendorf et~al.(2005)Altendorf, Restificar, and
  Dietterich]{Altendorf2005}
E.~E. Altendorf, A.~C. Restificar, and T.~G. Dietterich.
\newblock {Learning from Sparse Data by Exploiting Monotonicity Constraints}.
\newblock \emph{Proceedings of the Twenty-First Conference on Uncertainty in
  Artificial Intelligence (UAI2005)}, 2005.

\bibitem[Druzdzel and Henrion(1993)]{Druzdzel1993}
J.~Druzdzel and M.~Henrion.
\newblock {Efficient Reasoning in Qualitative Probabilistic Networks}.
\newblock In \emph{Proceedings of the Eleventh National Conference on
  Artificial Intelligence}, pages 548--553. AAAI Press, 1993.

\bibitem[Feelders and van~der Gaag(2005)]{Feelders2005}
A.~J. Feelders and L.~C. van~der Gaag.
\newblock {Learning Bayesian Network Parameters with Prior Knowledge about
  Context-Specific Qualitative Influences}.
\newblock \emph{Proceedings of the Twenty-First Conference on Uncertainty in
  Artificial Intelligence (UAI2005)}, 2005.

\bibitem[Gaag and de~Waal(2006)]{Gaag2006}
L.~C. Gaag and P.~de~Waal.
\newblock {Multi-dimensional Bayesian Network Classifiers.}
\newblock In \emph{Proceedings of the 3rd European Workshop on Probabilistic
  Graphical Models, PGM 2006}, pages 107--114, oct 2006.

\bibitem[Masegosa et~al.(2016)Masegosa, Feelders, and van~der
  Gaag]{Masegosa2016}
A.~R. Masegosa, A.~J. Feelders, and L.~van~der Gaag.
\newblock {Learning from incomplete data in Bayesian networks with qualitative
  influences}.
\newblock \emph{International Journal of Approximate Reasoning}, 69:\penalty0
  18--34, 2016.

\bibitem[Nielsen and Jensen(2007)]{Jensen2007}
T.~D. Nielsen and F.~V. Jensen.
\newblock \emph{{Bayesian Networks and Decision Graphs (Information Science and
  Statistics)}}.
\newblock Springer, 2007.

\bibitem[Olesen et~al.(1989)Olesen, Kjaerulff, Jensen, Jensen, Falck,
  Andreassen, and Andersen]{OLESEN1989}
K.~G. Olesen, U.~Kjaerulff, F.~Jensen, F.~V. Jensen, B.~Falck, S.~Andreassen,
  and S.~K. Andersen.
\newblock {A munin network for the median nerve-a case study on loops}.
\newblock \emph{Applied Artificial Intelligence}, 3\penalty0 (2-3):\penalty0
  385--403, jan 1989.

\bibitem[Pearl(1988)]{Pearl1988}
J.~Pearl.
\newblock \emph{{Probabilistic reasoning in intelligent systems: networks of
  plausible inference}}.
\newblock Morgan Kaufmann Publishers Inc., dec 1988.

\bibitem[Plajner and Vomlel(2016{\natexlab{a}})]{Plajner2016}
M.~Plajner and J.~Vomlel.
\newblock {Probabilistic Models for Computerized Adaptive Testing:
  Experiments}.
\newblock Technical report, ArXiv:, 2016{\natexlab{a}}.

\bibitem[Plajner and Vomlel(2016{\natexlab{b}})]{Plajner2016b}
M.~Plajner and J.~Vomlel.
\newblock {Student Skill Models in Adaptive Testing}.
\newblock In \emph{Proceedings of the Eighth International Conference on
  Probabilistic Graphical Models}, pages 403--414. JMLR.org,
  2016{\natexlab{b}}.

\bibitem[Plajner and Vomlel(2020)]{Plajner2020}
M.~Plajner and J.~Vomlel.
\newblock {Learning bipartite Bayesian networks under monotonicity
  restrictions}.
\newblock \emph{International Journal of General Systems}, 49\penalty0
  (1):\penalty0 88--111, 2020.

\bibitem[Rasch(1960)]{Rasch1960}
G.~Rasch.
\newblock \emph{{Studies in mathematical psychology: I. Probabilistic models
  for some intelligence and attainment tests.}}
\newblock Danmarks Paedagogiske Institut, 1960.

\bibitem[Restificar and Dietterich(2013)]{Restificar2013}
A.~C. Restificar and T.~G. Dietterich.
\newblock {Exploiting monotonicity via logistic regression in Bayesian network
  learning}.
\newblock Technical report, Corvallis, OR : Oregon State University, 2013.

\bibitem[Savicky and Vomlel(2007)]{Savicky2007}
P.~Savicky and J.~Vomlel.
\newblock {Exploiting tensor rank-one decomposition in probabilistic
  inference}.
\newblock \emph{Kybernetika}, 43\penalty0 (5):\penalty0 747--764, 2007.

\bibitem[van~der Gaag et~al.(2004)van~der Gaag, Bodlaender, and
  Feelders]{Gaag2004}
L.~van~der Gaag, H.~L. Bodlaender, and A.~J. Feelders.
\newblock {Monotonicity in Bayesian networks}.
\newblock \emph{20th Conference on Uncertainty in Artificial Intelligence (UAI
  '04)}, pages 569--576, 2004.

\bibitem[van~der Linden and Glas(2000)]{Linden2000}
W.~J. van~der Linden and C.~A.~W. Glas.
\newblock \emph{{Computerized Adaptive Testing: Theory and Practice}},
  volume~13.
\newblock Kluwer Academic Publishers, 2000.

\bibitem[Wellman(1990)]{Wellman1990}
M.~P. Wellman.
\newblock {Fundamental concepts of qualitative probabilistic networks}.
\newblock \emph{Artificial Intelligence}, 44\penalty0 (3):\penalty0 257--303,
  1990.

\end{thebibliography}
\end{document}